\documentclass{article}
\pdfoutput=1 
\PassOptionsToPackage{numbers, compress}{natbib}

\usepackage[preprint]{neurips_2024}
\usepackage[utf8]{inputenc} 
\usepackage[T1]{fontenc}    
\usepackage{hyperref}       
\usepackage{url}            
\usepackage{booktabs}       
\usepackage{amsfonts}       
\usepackage{nicefrac}       
\usepackage{microtype}      
\usepackage{xcolor}         
\usepackage{amsmath, amssymb}
\usepackage{graphicx}
\usepackage{wrapfig}
\graphicspath{ {./images/} }

\title{SignSpeak: Open-Source Time Series Classification for ASL Translation}

\author{
  Aditya Makkar\textsuperscript{*} \\
  Cheriton School of Computer Science \\
  University of Waterloo\\
  Waterloo, ON  \\
  \texttt{aditya.makkar@uwaterloo.ca} \\
  \And
  Divya Makkar\textsuperscript{*}\\
  Cheriton School of Computer Science \\
  University of Waterloo \\
  Waterloo, ON \\
  \texttt{divya.makkar@uwaterloo.ca} \\
  \And
  Aarav Patel\thanks{All authors contributed equally} \\
  Faculty of Engineering \\
  University of Waterloo \\
  Waterloo, ON \\
  \texttt{aarav.patel@uwaterloo.ca} \\
  \And
  Liam Hebert \\
  Cheriton School of Computer Science \\
  University of Waterloo \\ 
  Waterloo, ON \\ 
  \texttt{liam.hebert@uwaterloo.ca} \\
}

\begin{document}

\maketitle

\begin{abstract}

The lack of fluency in sign language remains a barrier to seamless communication for hearing and speech-impaired communities. In this work, we propose a low-cost, real-time ASL-to-speech translation glove and an exhaustive training dataset of sign language patterns. We then benchmarked this dataset with supervised learning models, such as LSTMs, GRUs and Transformers, where our best model achieved 92\% accuracy. The SignSpeak dataset has 7200 samples encompassing 36 classes (A-Z, 1-10) and aims to capture realistic signing patterns by using five low-cost flex sensors to measure finger positions at each time step at 36 Hz. Our open-source dataset, models and glove designs, provide an accurate and efficient ASL translator while maintaining cost-effectiveness, establishing a framework for future work to build on.  

\end{abstract}

\section{Introduction}
American Sign Language (ASL) is the most prominent sign language in North America \cite{nidcd_asl}, yet as of 2021, only 0.15\% of Americans are fluent in it \cite{kumar2021comparative}. This low figure causes significant challenges for hearing and speech-impaired individuals, including limited access to education, opportunities and essential services, leading to isolation and depression\cite{Verge_2023}. 

To address these barriers, prior work using optical-based methods has shown strong results in translating images of ASL gestures to speech; however, they are limited in real-world applicability\cite{10288799,s23125555}. CNN and vision-based transformer models necessitate using a camera pointed at a user's hands while signing, which is impractical in many contexts. Additionally, the use of cameras also presents a privacy risk by capturing the user and surrounding individuals while requiring considerable computing resources as frames must be sent to a server. This is infeasible and limits the scope of optical-based ASL translation within a real-world context. 

Sensor-based models using embedded devices have been introduced to treat ASL as a time-series multi-label classification problem to address the limitations of optical systems. However, many of these datasets are private \cite{electronics12081904,s20216256} and have not been trained on a well-practiced sign-based language such as ASL \cite{9616000}, limiting their applicability. To address this, we introduce SignSpeak, an open-source ASL dataset comprising of 7200 recordings of 36 classes. Our dataset was recorded using five low-cost flex sensors, one for each finger, for all letters and numbers in ASL. The scale of our dataset enables researchers to test novel models on a dataset collected in an environment aimed at real-world feasibility to progress ASL-to-speech efforts. We extensively benchmarked various methods on this dataset, with our best result achieving 92\% categorical accuracy, which matches or exceeds previous ASL time-series classification work \cite{article}.\footnote[1]{ The GitHub codebase and dataset are available at \\\texttt{\href{https://github.com/adityamakkar000/ASL-Sign-Research}{https://github.com/adityamakkar000/ASL-Sign-Research}} \\and \texttt{\href{https://doi.org/10.7910/DVN/ODY7GH}{https://doi.org/10.7910/DVN/ODY7GH}} respectively.}

\section{Related work}

Previous work using a glove-based apparatus involves sensory devices such as flex sensors and inertial measurement units (IMUs). \citet{electronics12081904} utilize flex sensor gloves to capture 37 hand gestures (numbers 0-10 and letters A-Z) using MLPs achieving 97.6\% accuracy. However, a fundamental flaw limits real-world applicability as the measurements are static and recorded at only one point during the gesture. This fails to account for ASL's dynamic nature since each sign is a sequence of motions that must be continuously measured. Furthermore, the dataset is closed-source, prohibiting others from building on it.  

\citet{s20216256} developed a glove taking continuous measurements of 6 inertial measurement units, including an accelerometer, gyroscope, and magnetometer. They report a 99.87\% accuracy; however, this study presents a drawback: each input is 10-15 seconds long and impractical for real-world signing, which is performed at a significantly faster rate of 4 syllables per second \cite{Wilbur_2009}. Similar to the previous work, the dataset was not released publicly.

\citet{s23125555} developed a 28-sensor glove which recorded 63 data channels to train an LSTM model.\footnote{Certain sensors measured multiple data channels. See the referenced paper for more details.} Multiple data channels and sensory equipment significantly increase the glove's cost, decreasing its affordability. In addition, less sensory equipment can produce similar results in commercial use. \citet{9616000} utilized a transformer architecture to achieve over 99\% accuracy on a synthetic glove-collected gesture dataset, preventing its applicability to the ASL community.  

We differ from previous work by introducing an open-source ASL dataset measuring 5 flex sensor channels. It includes 200 samples for all alphanumeric classes, allowing for a cost-effective and resource-efficient glove with broad applicability for the ASL community.   

\section{Methodology}

\begin{wrapfigure}{r}{0.3\textwidth} 
    \centering
    \includegraphics[width=0.3\textwidth]{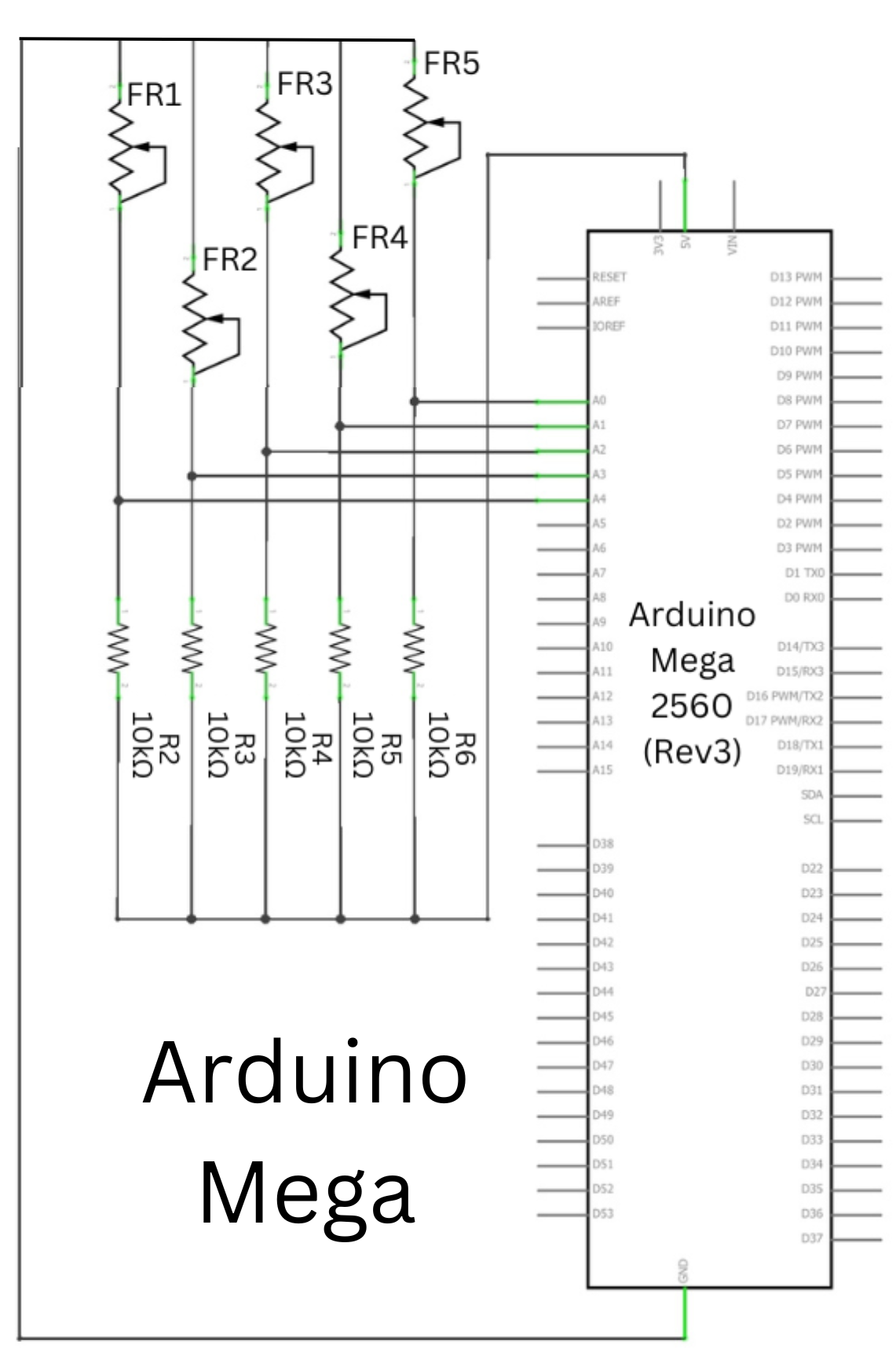}
    \caption{Circuit of data collection glove.}
    \label{fig:enter-label}
\end{wrapfigure}

\subsection{Data collection}

For this study, a glove was constructed with five parallel flex sensors on each finger in series with a 10,000$\Omega$ resistor. 5$V$ were applied and measured across each sensor with an Arduino MEGA 2560. We recorded each feature within the standard Arduino 10-bit range of [0, 1023]. Each gesture was recorded at 36 Hz while ensuring that the sum of all flex sensor measurements was below 5000 or  24.4$V$. This value was experimentally determined, and indicates that the fingers were flexed (the sign being performed), allowing for intentional data collection. We retain all gesture recordings between 1.38 and 2.22 seconds (50-time to 80-time) steps to ensure that accidental gestures were not added and that the gestures reflect realistic signing patterns. 

\subsection{Model architecture}

Each gesture recording contains $C=5$ channels and has a maximum time dimension of $T=79$ with all input features 0-padded to ensure a consistent batch size.  We benchmarked RNN and Transformer-based time series models on the SignSpeak dataset. In particular, we evaluated a 2-layer LSTM\cite{10.1162/neco.1997.9.8.1735} and a 2-layer GRU\cite{DBLP:journals/corr/ChoMGBSB14} model, where the output from the last cell was fed into a 2-layer MLP Softmax classification layer. We then applied a dropout layer to reduce overfitting\cite{DBLP:journals/corr/abs-1207-0580}. 

\begin{equation}
\label{eq:lstm}
    \mathbf{y_0} = \text{LSTM}(\mathbf{x})
\end{equation}
\begin{equation}
\label{eq:lstm2}
    \mathbf{y_{\textbf{output}}} = \text{SOFTMAX}(\text{MLP}(\text{MLP}(\text{LSTM}(\mathbf{y_0})^{(T)})))
\end{equation}

For the Stacked LSTM model, refer to eq. \ref{eq:lstm} and \ref{eq:lstm2} where $ \mathbf{x} \in \mathbb{R}^{T \times C} \text{and } \mathbf{y_{output}^{\top} \in \mathbb{R}^{ \text{classes} }}$.  \citet{article} presented a dense-LSTM network for EMG-ASL classification, showing that a MLP projection before the RNN could achieve state-of-the-art results. This work used a 2-layer MLP Softmax classifier following the dense-RNN unit. 

\begin{equation}
\label{eq:rnn}
    \mathbf{y_0} = [MLP(\mathbf{x}^{(0)}), MLP(\mathbf{x}^{(1)}), \ldots, MLP(\mathbf{x}^{(T)})]
\end{equation}
\begin{equation}
\label{eq:rnn2}
    \mathbf{y_{\textbf{output}}} = \text{SOFTMAX}(\text{MLP}(\text{MLP}(\text{LSTM}(\mathbf{y_0})^{(T)})))
\end{equation}

Eq. \ref{eq:rnn} and 
\ref{eq:rnn2} describe the dense-LSTM. For a dense-stacked RNN, eq. \ref{eq:rnn2} is modified by composing the RNN function with itself for the input $\mathbf{y_0}$. Each RNN gate used a Sigmoid activation, while MLPs used a Tanh activation. The hidden size of the RNN cells was $h_{\text{RNN}}=64$, and the dense and/or output MLP was $h_{\text{MLP}}=128$. 

In recent literature, transformers have matched or exceeded SOTA benchmarks in time-series classification. \citet{9616000} WaveGlove Encoder, based on Transformers \cite{DBLP:journals/corr/VaswaniSPUJGKP17}, have surpassed previous SOTA architectures on this task \cite{9616000}. Inspired by this architecture, we benchmark a slightly modified version of WaveGlove on SignSpeak, adding a classification token ([CLS]) to the start of the input as done with BERT\cite{DBLP:journals/corr/abs-1810-04805}. The input is passed through a learnable embedding and positional embedding table with the projected input being fed into an Encoder \cite{DBLP:journals/corr/VaswaniSPUJGKP17} with layer normalization before the self-attention and MLPs, as described by \citet{DBLP:journals/corr/abs-2010-11929}. The input format $\mathbf{x} \in \mathbb{R}^{T \times C}$ represents 5-flex sensor channels across time $T$, before being projected into a dimension $D =32$ with the sequential nature encoded by the positional embedding. We utilized the GELU activation function\cite{DBLP:journals/corr/HendrycksG16} and the number of layers was $L = 5$. All Encoder and RNN parameters were found through a Cartesian product hyperparameter sweep. 

\begin{equation}
    \mathbf{y_0} = [\mathbf{x}^{\text{class}}, \mathbf{x}\mathbf{E}_{\text{emb}}] + \mathbf{E}_{\text{pos\_emb}}
\end{equation}
\begin{equation}
    \mathbf{y_{l}} = \text{Encoder}(\mathbf{y_l}), \quad \text{where } l=1,2,\ldots, L
\end{equation}
\begin{equation}
    \mathbf{y_{\text{output}}} = \text{SOFTMAX}(\text{MLP}(\text{LN}(\mathbf{y_L}^{(0)}))
\end{equation}

The encoder is described by eq.(5) - (7), where $ \mathbf{E}_{\text{emb}} \in \mathbb{R}^{C \times D}, \ \mathbf{E}_{\text{pos\_emb}} \in \mathbb{R}^{(T+1) \times D}$.

\section{Results} 

\begin{table}[]
    \centering
    \caption{Model description and results on accuracy, F1-score, $\sigma_{accuracy}$ and $\sigma_{F1 Score}$ }
    \begin{tabular}{c c c c c c}
     \hline
        Model & Parameters & Categorical Accuracy & F1Score & $\sigma_{accuracy}$ & $\sigma_{F1Score}$\\ 
        \hline\hline
        Dense LSTM  & 63K & 0.8348	& 0.8301 & 0.0110 &0.0100 \\
         \hline
        Dense GRU  & 51K & 0.6692 & 0.6574 & 0.0565 & 0.0631  \\
         \hline
       Stacked LSTM & 64K & 0.9167	& 0.9164 & 0.0067	& 0.0068  \\
         \hline
        Stacked GRU & 51K & 0.9221	& 0.9218 & 0.0109	& 0.0110  \\
         \hline
        Dense Stacked LSTM & 96K & 0.8876 & 0.8873 & 0.0194 & 0.0196  \\
         \hline
        Dense Stacked GRU & 76K  & 0.9192	& 0.9188 & 0.0079	& 0.0079  \\
         \hline
         Encoder & 67K &  0.9136 & 0.8873 & 0.0078 & 0.0195\\
         \hline
    \end{tabular}
    \label{tab:results}
\end{table}
All models were trained with AdamW \cite{DBLP:journals/corr/abs-1711-05101}, with $B_1 = 0.9, B_2 = 0.999$, a weight decay of 0.01, and a plateau learning rate decay on validation loss with a patience of 20 epochs of 0.5 starting from 0.001 until a minimum learning rate of 0.0001. RNNs were trained with a batch size of 64 for 15 minutes on an M2, and the Encoder was trained with a batch size of 256 for 15 minutes on a T4 GPU. All models used a 0.2 dropout probability. The metrics used to evaluate all models were categorical accuracy and the F1-Score. We utilized a stratified 5-fold validation and reported the standard deviation and average result of the held-out folds.

The results in Table \ref{tab:results} indicate methods on private datasets do not generalize to the SignSpeak dataset. This may be due to a reduction in the number of data channels. This can be seen with a model such as the Transformer where a lack of data channels reduces its performance. In particular, we found that simple models such as a stacked GRU perform the best, whereas models such as a dense LSTM proposed by \citet{article} do not achieve near state-of-the-art results.  We believe that the potential for Transformer-based architectures can be unlocked with more training data, which can then be further fine-tuned on the SignSpeak dataset. Our leading RNN and Encoder models still maintain above 99.5\% traditional accuracy, demonstrating performance on par with previous studies. 

\begin{wrapfigure}{r}{0.5\textwidth} 
    \centering
    \includegraphics[width=0.5\textwidth]{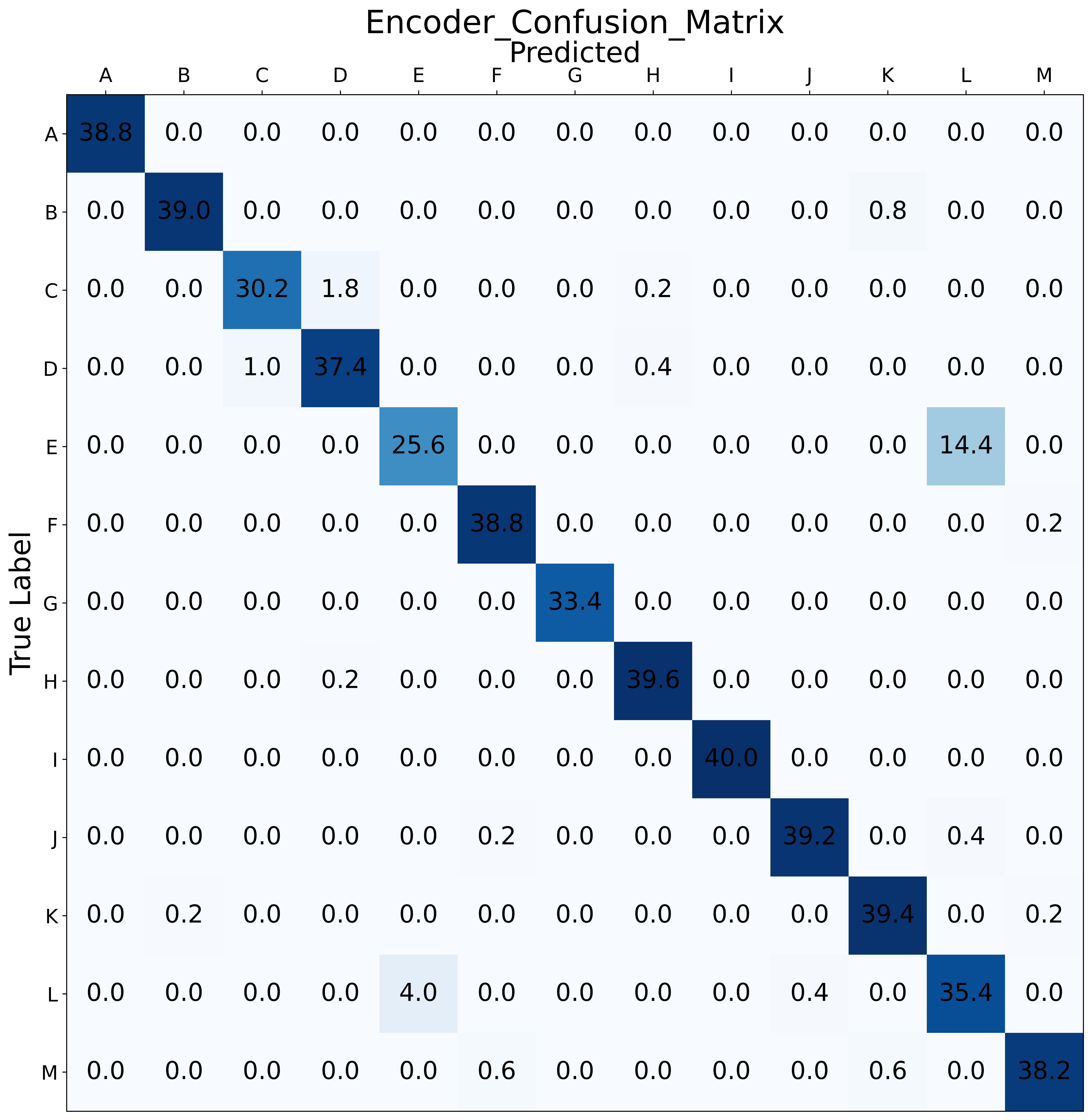}
    \caption{Confusion matrix for the Encoder classes of 'A' to 'M' to highlight the poorest performance.}
    \label{fig:confusion-matrix}
\end{wrapfigure}

Additionally, the Transformer architecture has the largest difference between F1-score and accuracy, indicating a bias towards certain classes. Figure \ref{fig:confusion-matrix} displays the confusion matrix and it is evident the low accuracy is due to specific classes such as 'E' and 'L'. Specifically, the Encoder incorrectly predicts 'L' 36\% of the time when the actual label is 'E'. Additionally in ASL, these letters do not share the same features; 'E' is very similar to a letter such as 'A'. This indicates the model's over-predicts between certain classes and could be outlier patterns in the dataset. Analyzing this class with stacked GRU and LSTM models, they predict 'L' instead of 'E' 16\% and 13\% of the time, respectively. This indicates it is a learned bias of an Encoder model but over-fitting is still present in all models. 

\section{Future work }

The models presented in this study only required a moderate amount of computing power to achieve 92\% accuracy. In the future, leveraging more powerful computing resources can enable the implementation of larger-scale architectures to further enhance performance. Additionally, the gestures chosen in this dataset (alphanumeric classes) reflect an extremely limited subsection of real-world ASL; thus, future work is aimed at expanding the dataset by collecting data and creating classes for phrases and actions that resemble daily communication making the product viable for commercial use. Lastly, while our measurements were recorded at 36 Hz, which is slower than average ASL signing rates, we anticipate that using an improved MCU will allow us to increase this frequency to 200 Hz, aligning with more realistic signing speeds \cite{Wilbur_2009}. These advancements will expand on our existing research and contribute to a more refined product that can facilitate the integration of hearing and speech-impaired individuals into society.

\section{Conclusion}

In this study, we presented SignSpeak, an open-source dataset collected using a custom low-cost glove architecture benchmarked on time-series classification models to mimic real-time ASL translation. We found that a stacked GRU achieves the strongest results on categorical accuracy. SignSpeak benefits speech and hearing-impaired communities by providing a way to benchmark models on a universal dataset. Our work on Signspeak can provide a foundation for researchers to build upon our open-source dataset, leveraging supervised learning techniques to deliver assistive and accessible technology to communities in need.

\bibliography{neurips_2024}

\input

\end{document}